\def\year{2018}\relax
\documentclass{article}
\usepackage{iclr2018_conference,times}
\def\iclrfinal

\usepackage{natbib}

\usepackage[colorlinks=true,allcolors=blue!50!black]{hyperref}
\usepackage{url}
\usepackage{amsfonts,amsmath,algorithmic,algorithm}
\usepackage{bm}
\usepackage{tikz}
\usetikzlibrary{bayesnet}
\usepackage{xspace}
\usepackage{wrapfig}
\usepackage{graphicx}
\usepackage{authblk}

\RequirePackage[textsize=scriptsize]{todonotes}
\RequirePackage[skip=1pt,font=small,justification=centering]{caption}
\usepackage{subcaption}
\makeatletter
\let\c@table\c@figure
\let\ftype@table\ftype@figure
\makeatother

\RequirePackage{enumitem}
\setlist{nolistsep}
\usepackage{titlesec}
\titlespacing*{\paragraph}{0pt}{0ex plus .1ex}{1em}

\belowdisplayskip \abovedisplayskip
\definecolor{turquoise}{cmyk}{0.65,0,0.1,0.1}
\definecolor{purple}{rgb}{0.65,0,0.65}
\definecolor{dark_green}{rgb}{0, 0.5, 0}
\definecolor{orange}{rgb}{0.8, 0.6, 0.2}
\definecolor{brown}{rgb}{0.5, 0.16, 0.16}

\newcommand{\shiv}[1]{{\color{brown}[Shiv: #1]}}

\title{\raggedright
Generalizing Across Domains via Cross-Gradient Training
}
\newcommand*\samethanks[1][\value{footnote}]{\footnotemark[#1]}

\author[1]{Shiv Shankar\thanks{These two authors contributed equally}}
\author[1] {Vihari Piratla\samethanks }
\author[1]{Soumen Chakrabarti}
\author[1,2]{Siddhartha Chaudhuri}
\author[1]{Preethi Jyothi}
\author[1]{Sunita Sarawagi}
\affil[1]{Department of Computer Science, Indian Institute of Technology Bombay}
\affil[2]{Adobe Research}

\def\jac{\mathbb{J}}
\def\R{\mathbb{R}}
\newcommand{\vek}[1]{\ensuremath{\text{\bf {#1}}}}
\newcommand{\vx}{{\vek{x}}}

\newcommand{\vg}{{\vek{g}}}
\newcommand{\vgmap}{\hat{\vek{g}}}
\newcommand{\vgmapcoord}{\hat{g}}

\newcommand{\cD}{{\mathcal{D}}}
\newcommand{\cY}{{\mathcal{Y}}}

\newcommand{\argmax}{{\text{argmax}}}

\newcommand{\dan}{\textsc{DAN}\xspace}
\newcommand{\goodfellow}{\textsc{LabelGrad}\xspace}
\newcommand{\crossgrad}{\textsc{CrossGrad}\xspace}
\newcommand{\mypara}[1]{\medskip\noindent\textbf{#1}~}

\iclrfinalcopy
\begin{document}
%\title{\sharedtitle}
\maketitle

\begin{abstract}
We present \crossgrad, a method to use multi-domain training data to learn a  classifier that generalizes to new domains.  \crossgrad\ does not need an adaptation phase via labeled or unlabeled data, or domain features in the new domain. Most existing domain adaptation methods attempt to erase domain signals using techniques like domain adversarial training.  In contrast, \crossgrad\ is free to use domain signals for predicting labels, if it can prevent overfitting on training domains.  We conceptualize the task in a Bayesian setting, in which a sampling step is implemented as data augmentation, based on domain-guided perturbations of input instances.  \crossgrad\ parallelly trains a label and a domain classifier on examples perturbed by loss gradients of each other's objectives.  This enables us to directly perturb inputs, without separating and re-mixing domain signals while making various distributional assumptions.  Empirical evaluation on three different applications where this setting is natural establishes that (1)~domain-guided perturbation provides consistently better generalization to unseen domains, compared to generic instance perturbation methods, and that (2)~data augmentation is a more stable and accurate method than domain adversarial training.\footnote{Code and dataset can be found at \protect\url{https://github.com/vihari/crossgrad}}
\end{abstract}

\section{Introduction}
\label{sec:Intro}

We investigate how to train a classification model using multi-domain training data, so as to generalize to labeling instances from unseen domains.  This problem arises in many applications, viz., handwriting recognition, speech recognition, sentiment analysis, and sensor data interpretation.  In these applications, domains may be defined by fonts, speakers, writers, etc.  Most existing work on handling a target domain not seen at training time requires either labeled or unlabeled data from the target domain at test time.  Often, a separate ``adaptation'' step is then run over the source and target domain instances, only after which target domain instances are labeled.  In contrast, we consider the situation where, during training, we have labeled instances from several domains which we can collectively exploit so that the trained system can handle new domains without the adaptation step.

\subsection{Problem statement}

Let $\cD$ be a space of domains. During training we get labeled data from a proper subset $D \subset \cD$ of these domains.  Each labeled example during training is a triple $(\vx, y, d)$ where $\vx$ is the input, $y \in \cY$ is the true class label from a finite set of labels $\cY$ and $d \in D$ is the domain from which this example is sampled.  We must train a classifier to predict the label $y$ for examples sampled from all domains, including the subset $\cD \setminus D$ not seen in the training set.  Our goal is high accuracy for both in-domain (i.e., in $D$) and out-of-domain  (i.e., in $\mathcal{D}\setminus D$) test instances.

% During deployment we do not get any domain label on an unlabeled example $\vx$ for which the predicted label is sought.  Also, unlike in prior work we do not assume a separate model adaptation phase  on the target domain using a labeled dataset~\cite{SantoroBBWL16,VinyalsBLKW16,Finn2017ModelAgnosticMF,ravi17} or an unlabeled dataset~\cite{zhao17,tzeng15,ganin16}.  Neither do we assume the availability of task-irrelevant training pairs from fixed source and target domains \sunita{Zero-shot wrongly used?}\cite{PengWE17}.

% An implicit assumption in this task is that $\Pr(y|\vx, d)=\Pr(y|\vx)$.  That is, any dependence of the label on the domain is transferred via the input $\vx$. However, we do not assume that $\Pr(\vx|d) = \Pr(\vx)$. For generalization across domains $\Pr(\vx|d)$ of different domains have to be related.  We provide a probabilistic interpretation of these assumptions in Section~\ref{sec:probmodel}.

One challenge in learning a classifier that generalizes to unseen domains is that $\Pr(y|\vx)$ is typically harder to learn than $\Pr(y|\vx,d)$.
While \cite{Yang2015} addressed a similar setting, they assumed a specific geometry characterizing the domain, and performed kernel regression in this space.  In contrast, in our setting, we wish to avoid any such explicit domain representation, appealing instead to the power of deep networks to discover implicit features.

Lacking any feature-space characterization of the domain, conventional training objectives (given a choice of hypotheses having sufficient capacity) will tend to settle to solutions that overfit on the set of domains seen during training.  A popular technique in the domain adaptation literature to generalize to new domains is domain adversarial training~\citep{ganin16,tzeng17}.  As the name suggests, here the goal is to learn a transformation of input \vx\ to a domain-independent representation, with the hope that amputating domain signals will make the system robust to new domains.  We show in this paper that such training does not necessarily safeguard against over-fitting of the network as a whole.  We also argue that even if such such overfitting could be avoided, we do not necessarily want to wipe out domain signals, if it helps in-domain test instances.

\subsection{Contribution}

In a marked departure from domain adaptation via amputation of domain signals, we approach the problem using a form of data augmentation based on domain-guided perturbations of input instances.  If we could model exactly how domain signals for $d$ manifest in \vx, we could simply replace these signals with those from suitably sampled other domains $d'$ to perform data augmentation.  We first conceptualize this in a Bayesian setting: discrete domain $d$ `causes' continuous multivariate \vg, which, in combination with $y$, `causes' \vx.  Given an instance \vx, if we can recover \vg, we can then perturb \vg\ to $\vg'$, thus generating an augmented instance $\vx'$.  Because such perfect domain perturbation is not possible in reality, we first design an (imperfect) domain classifier network to be trained with a suitable loss function.  Given an instance \vx, we use the loss gradient w.r.t. \vx\ to perturb \vx\ in directions that change the domain classifier loss the most.  The training loss for the $y$-predictor network on original instances is combined with the training loss on the augmented instances.  We call this approach \emph{cross-gradient training}, which is embodied in a system we describe here, called \crossgrad.   We carefully study the performance of \crossgrad\ on a variety of domain adaptive tasks: character recognition, handwriting recognition and spoken word recognition.  We demonstrate performance gains on new domains without any out-of-domain instances available at training time.

% A recent approach for domain classification is to create an internal representation that is invariant to domain changes.   Hard task.  Not required.  Simpler to exploit multiple domains and handle local perturbations.

\section{Related Work}
Domain adaptation has been studied under many different settings: two domains~\citep{ganin16,tzeng17} or multiple domains~\citep{Mansour2009,Zhang2015}, with target domain data that is labeled~\citep{Daume2007,Kumar2010,Saenko2010} or unlabeled~\citep{Gopalan2011,Gong2012,ganin16}, paired examples from source and target domain~\citep{PengWE17}, or domain features attached with each domain~\citep{Yang2016}. Domain adaptation techniques have been applied to numerous tasks in speech, language processing and computer vision \citep{Woodland2001,Saon2013,Jiang2007,Daume2007,Saenko2010,Gopalan2011,Li13,HuangB17,UpchurchSB16}. However, unlike in our setting, these approaches typically assume the availability of some target domain data which is either labeled or unlabeled.

% \cite{UpchurchSB16} extract a style vector for each training style using an adversarial content classifier.  A generator takes as input the extracted style vector and a content label to generate an image of that content in the given style.  Our goal is different.

For neural networks a recent popular technique is domain adversarial networks (DANs)~\citep{tzeng15,tzeng17,ganin16}.  The main idea of DANs is to learn a representation in the last hidden layer (of a multilayer network) that cannot discriminate among different domains in the input to the first layer.  A domain classifier is created with the last layer as input.  If the last layer encapsulates no domain information apart from what can be inferred from the label, the accuracy of the domain classifier is low.  The DAN approach makes sense when all domains are visible during training.  In this paper, our goal is to generalize to unseen domains.

%\shiv{Please review the next three paras -- They are added new}
Domain generalization is traditionally addressed by learning representations that encompass information from all the training domains.
\cite{MuandetBS13} learn a kernel-based representation that minimizes domain dissimilarity and retains the functional relationship with the label.
\cite{GanYG16} extends \cite{MuandetBS13} by exploiting attribute annotations of examples to learn new feature representations for the task of attribute detection.
In \cite{GhifaryBZB15}, features that are shared across several domains are estimated by jointly learning multiple data-reconstruction tasks.
Such representations are shown to be effective for domain generalization, but ignore any additional information that domain features can provide about labels.

% However, the method is not relevant in our case because of the lack of any useful attribute annotations for the examples.

%The method proposed in \cite{Da Li 18} also relies on ignoring or erasing the domain dependent features for the domain generalization.

%Apart from the fact that it is computationally hard to train an autoencoder to reconstruct large inputs, this method extracts features shared across domains.
%% This work in essence is similar to DAN/\cite{Ganin16}
Domain adversarial networks (DANs)~\citep{ganin16} can also be used for domain generalization in order to learn domain independent representations.
A limitation of DANs is that they can be misled by a representation layer that over-fits to the set of training domains.
In the extreme case, a representation that simply outputs label logits via a last linear layer (making the softmax layer irrelevant) can keep both the adversarial loss and label loss small, and yet not be able to generalize to new test domains.
In other words, not being able to infer the domain from the last layer does not imply that the classification is domain-robust.

Since we do not assume any extra information about the test domains, conventional approaches for regularization and generalizability are also relevant. \cite{XuLNX14} use exemplar-based SVM classifiers regularized by a low-rank constraint over predictions. \cite{ECCV12_Khosla} also deploy SVM based classifier and regularize the domain specific components of the learners.
%Loss on meta-data is used as a regularizer in \cite{dali}.
The method most related to us is the adversarial training of ~\citep{szegedy14,goodfellow2014,MiyatoDG16} where examples perturbed along the gradient of classifier loss are used to augment the training data. perturbs examples. Instead, our method attempts to model domain variation in a continuous space and perturbs examples along domain loss.
% There also exist approaches that employ regularization to learn domain-robust models. \cite{XuLNX14} use exemplar-based SVM classifiers regularized by a low-rank constraint over classifier predictions.
% \citeauthor{zhao17} proposes to make the domain-classifier conditional on labels too.

Our Bayesian model to capture the dependence among label, domains, and input is similar to \citet[Fig.~1d]{Zhang2015}, but the crucial difference is the way the dependence is modeled and estimated. Our method attempts to model domain variation in a continuous space and project perturbation in that space to the instances.

% domain features in the target domain, Other auxillary aids to domain adaptation:
% \cite{PengWE17} relies on example pairs from source and target domain from a irrelevant label space.  This is called zero-shot domain adaptation

% Todo: read this paper: https://arxiv.org/abs/1704.03453 \\
%\pj{There's a lot of prior work on multi-source domain adaptation (starting with~\cite{Mansour2009} to the more recent~\cite{Zhang2015}) which assumes the same setting we do. Should we cite some of these papers?}

\section{Our approach}
\label{sec:probmodel}

We assume that input objects are characterized by two uncorrelated or weakly correlated tags: their {\em label} and their {\em domain}. E.g. for a set of typeset characters, the label could be the corresponding character of the alphabet (`A', `B' etc) and the domain could be the font used to draw the character.   In general, it should be possible to change any one of these, while holding the other fixed.

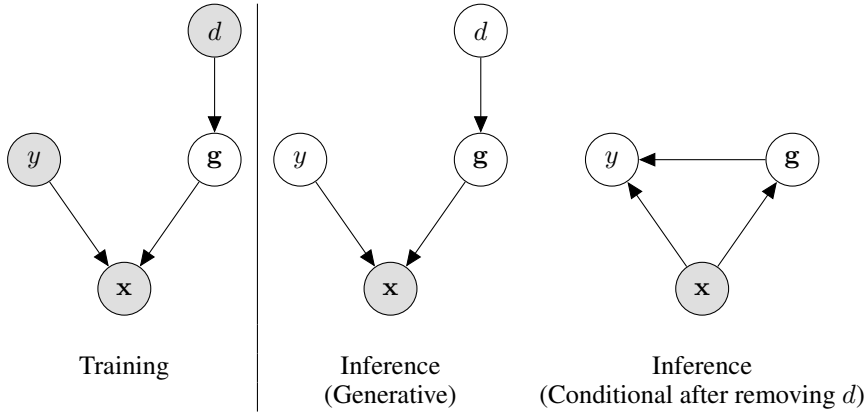
\begin{figure}[ht]
  \begin{center}
  \begin{tabular}{c|cc}
\begin{tikzpicture}
  % Define nodes
  \node[obs]                               (x) {$\mathbf{x}$};
% \node[latent, above=of x, xshift=-1.2cm] (f) {$\mathbf{f}$};
  \node[latent, above=of x, xshift=1.2cm]  (g) {$\mathbf{g}$};
  \node[obs, above=of x, xshift=-1.2cm]  (y) {$y$};
  \node[obs, above=of g]            (d) {$d$};
  % Connect the nodes
  \edge {g, y} {x} ; %
%  \edge {y} {f} ;
  \edge {d} {g} ;
  % Plates
%  \plate {yx} {(x)(y)} {$N$} ;
%  \plate {} {(w)(y)(yx.north west)(yx.south west)} {$M$} ;
\end{tikzpicture}
&
\begin{tikzpicture}
  % Define nodes
  \node[obs]                               (x) {$\mathbf{x}$};
% \node[latent, above=of x, xshift=-1.2cm] (f) {$\mathbf{f}$};
  \node[latent, above=of x, xshift=1.2cm]  (g) {$\mathbf{g}$};
  \node[latent, above=of x, xshift=-1.2cm]  (y) {$y$};
  \node[latent, above=of g]            (d) {$d$};
  % Connect the nodes
  \edge {g, y} {x} ; %
%  \edge {y} {f} ;
  \edge {d} {g} ;
\end{tikzpicture}
&
\begin{tikzpicture}
  % Define nodes
  \node[obs]                               (x) {$\mathbf{x}$};
% \node[latent, above=of x, xshift=-1.2cm] (f) {$\mathbf{f}$};
  \node[latent, above=of x, xshift=1.2cm]  (g) {$\mathbf{g}$};
  \node[latent, above=of x, xshift=-1.2cm]  (y) {$y$};
  \edge {x} {g, y} ; %
%  \edge {y} {f} ;
  \edge {g} {y} ;
\end{tikzpicture}\\
& \\
Training & Inference & Inference \\
 & (Generative) & (Conditional after removing $d$)
\end{tabular}
\end{center}
  \caption{\label{fig:bn} Bayesian network to model dependence between label $y$, domain $d$, and input $\vx$.}
\end{figure}

We use a Bayesian network to model the dependence among the label $y$, domain $d$, and input $\vx$ as shown in Figure~\ref{fig:bn}.  Variables $y \in \cY$ and $d \in \cD$ are discrete and $\vg \in \R^q,\vx \in \R^r$ lie in continuous multi-dimensional spaces.
% The label $y$ induces a set of latent content features $\vf$  and
The domain $d$ induces a set of latent domain features $\vg$.  The input $\vx$ is obtained by a complicated, un-observed mixing\footnote{The dependence of $y$ on $\vx$ could also be via continuous hidden variables but our model for domain generalization is agnostic of such structure.} of $y$ and $\vg$. In the training sample $L$, nodes $y,d,\vx$ are observed but $L$ spans only a proper subset $D$ of the set of all domains $\cD$.  During inference, only $\vx$ is observed and we need to compute the posterior $\Pr(y|\vx)$.
As reflected in the network, $y$ is not independent of $d$ given $\vx$.  However, since $d$ is discrete and we observe only a subset of $d$'s during training, we need to make additional assumptions to ensure that we can generalize to a new $d$ during testing.
The assumption we make is that integrated over the training domains the distribution $P(\vg)$ of the domain features is well-supported in $L$.  More precisely, generalizability of a training set of domains $D$ to the universe of domains $\cD$ requires that
\begin{align}
\label{eq:dg}
P(\vg) &= \sum_{d \in \cD} P(\vg | d) P(d)
\approx \sum_{d \in D} P(\vg | d) \frac{P(d)}{P(D)} \tag{A1}
\end{align}
Under this assumption $P(\vg)$ can be modeled during training, so that during inference we can infer $y$ for a given $\vx$ by estimating
\begin{equation}
\Pr(y|\vx) = \sum_{d\in \cD}\Pr(y|\vx,d)\Pr(d|\vx) = \int_\vg \Pr(y|\vx,\vg)\Pr(\vg|\vx) \approx \Pr(y|\vx,\hat{\vg})
\end{equation}
where $\hat{\vg} = \argmax_\vg \Pr(\vg|\vx)$ is the inferred continuous representation of the domain of $\vx$. 

This assumption is key to our being able to claim generalization to new domains even though most real-life domains are discrete.  For example, domains like fonts and speakers are discrete, but their variation can be captured via latent continuous features (e.g. slant, ligature size etc. of fonts; speaking rate, pitch, intensity, etc. for speech).  The assumption states that as long as the training domains span the latent continuous features we can generalize to new fonts and speakers. 

We next elaborate on how we estimate $\Pr(y|\vx,\hat{\vg})$ and $\hat{\vg}$ using the domain labeled data \mbox{$L=\{(\vx, y, d)\}$}.
% This assumption by itself does not guarantee that standard cross-entropy training of $\Pr(y|\vx)$ as a neural network on $L$ will generalize, because in many applications the $\Pr(y|\vx,d)$ is significantly easier to train.  So the network might just generate separate classifiers for each distinct training domain.  We must encourage the network to stay away from such easy solutions.
The main challenge in this task is to ensure that the model for $\Pr(y|\vx,\hat{\vg})$ is not over-fitted on the inferred $\vg$'s of the training domains. In many applications, the per-domain $\Pr(y|\vx,d)$ is significantly easier to train.  So, an easy local minima is to choose a different $\vg$ for each training $d$ and generate separate classifiers for each distinct training domain.  We must encourage the network to stay away from such easy solutions.
We strive for generalization by moving along the continuous space $\vg$ of domains to sample new training examples from hallucinated domains.
% However, a critical difference from much recent work is that we do this not by explicit distributional assumptions coupling $\vg$ and $\vx$, but instead via a network that is being trained.
Ideally, for each training instance $(\vx_i, y_i)$ from a given domain $d_i$, we wish to generate a new $\vx'$ by transforming its (inferred) domain $\vg_i$ to a random domain sampled from
% \sunita{$\Pr(\vg)$ or $P(\vg|y_i)$? No, this is p(g)}
$P(\vg)$, keeping its label $y_i$ unchanged.   Under the domain continuity assumption \eqref{eq:dg}, a model trained with such an ideally augmented dataset is expected to generalize to domains in $\cD\setminus D$.

However, there are many challenges to achieving such ideal augmentation.  To avoid changing $y_i$, it is convenient to draw a sample $\vg$ by perturbing $\vg_i$.  But $\vg_i$ may not be reliably inferred, leading to a distorted sample of $\vg$.   For example, if the $\vg_i$ obtained from an imperfect extraction conceals label information, then big jumps in the approximate $\vg$ space could change the label too.  We propose a more cautious data augmentation strategy that perturbs the input to make only small moves along the estimated domain features, while changing the label as little as possible.  We arrive at our method as follows.

\begin{wrapfigure}[15]{r}[25pt]{0.45\textwidth}
  \centering
  \vspace{-10mm}
  \includegraphics[scale=.4]{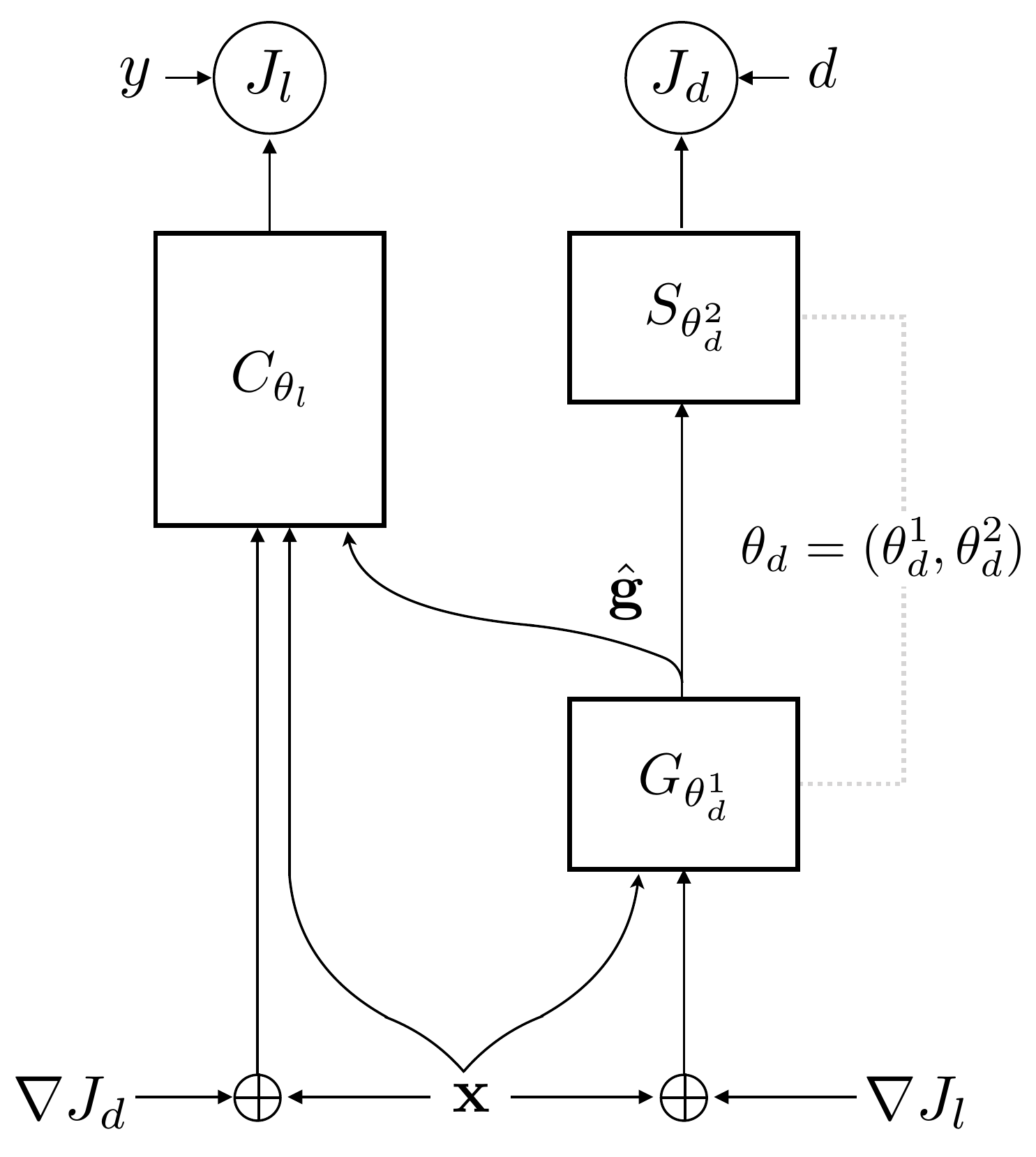}
  \caption{{\crossgrad} network design.}
  \label{fig:network}
\end{wrapfigure}

\mypara{Domain inference.}
We create a model $G(\vx)$ to extract domain features $\vg$ from an input $\vx$.  %We supervise the training of $G$ by applying softmax on $G(\vx_i)$ to predict the domain label $d_i$.
We supervise the training of $G$ to predict the domain label $d_i$ as $S(G(\vx_i))$ where $S$ is a softmax transformation.
We use $J_{d}$ to denote the cross-entropy loss function of this classifier.  Specifically, %\todo{Use $J_\text{dom}$ instead of potentially ambiguous subscript $d$?}
$J_d(\vx_i,d_i)$ is the domain loss at the current instance.

\mypara{Domain perturbation.}
Given an example $(\vx_i,y_i,d_i)$, we seek to sample a new example $(\vx'_i,y_i)$ (i.e., with the same label $y_i$), whose domain is as ``far'' from $d_i$ as possible. To this end, consider setting $\vx'_i = \vx_i + \epsilon \nabla_{\vx_i} J_d(\vx_i, d_i)$. Intuitively, this perturbs the input along the direction of greatest domain change\footnote{We use $\nabla_{\vx_i} J_d$ as shorthand for the gradient $\nabla_{\vx} J_d$ evaluated at $\vx_i$.}, for a given budget of $||\vx'_i-\vx_i||$. However, this approach presupposes that the direction of domain change in our domain classifier is %\todo{Could another way be to retain components of $\nabla_{\hat{\vg}}J_{d}$ that are perpendicular to $\nabla_{\hat{\vg}}J_\ell$?}
not highly correlated with the direction of label change. To enforce this in our model, we shall train the domain feature extractor $G$ to avoid domain shifts when the data is perturbed to cause label shifts.

What is the consequent change of the continuous domain features $\vgmap_i$? This turns out to be $\epsilon \jac \jac^\top \nabla_{\vgmap_i} J_d(\vx_i,d_i)$, where $\jac$ is the Jacobian of $\vgmap$ w.r.t. $\vx$. Geometrically, the $\jac \jac^\top$ term is the (transpose of the) metric tensor matrix accounting for the distortion in mapping from the $\vx$-manifold to the $\vgmap$-manifold. While this perturbation is not very intuitive in terms of the direct relation between $\vgmap$ and $d$, we show in the Appendix that the input perturbation \mbox{$ \epsilon \nabla_{\vx_i} J_d(\vx_i,d_i)$} is {\em also} the first step of a gradient descent process to induce the ``natural'' domain perturbation \mbox{$\Delta \vgmap_i = \epsilon' \nabla_{\vgmap_i} J_d(\vx_i,d_i)$}.

%Now for transforming the domain of an example $(x_i,y_i)$ from domain $d_i$, we sample a $\vg'$ near the estimated domain features $\vgmap_i = G(\vx_i)$ as $\vgmap_i + \epsilon \nabla_g J_d(d_i)$.
%\todo{Elaborate and justify}.  We attempt to safeguard against possible label shifts by training the $J_d$ loss symmetrically via an augmented dataset obtained by perturbing labels.

%
% require us to to estimate these three functions:
% \begin{enumerate}
% \item a function $G(\vx)$ that can recover the latent domain features $\vg_i$ from a given input $\vx_i$
% \item a function for $\Pr(\vg)$ from which we can sample a $\vg' \ne \vg_i$
% \item a function to transform the domain of $\vx_i$ to $\vg'$ $F(\vx_i, \vg_i \mapsto \vg', y_i)$ while keeping the label unchanged.
% \end{enumerate}
% If we knew true values of the above three functions, then the above procedure will generate an augmented dataset that is representative for all domains.  With imperfect estimated functions, we need to follow a more cautious method. For example, if the $\vg_i$ obtained from an imperfect $G(\vx)$ contains label information, then big jumps in the approximate $\vg$  space could change the label too.

% If we could reverse estimate g perfectly given x, and at least for a fixed y we knew how to generate a x' from sampled g' then the obvious data augmentation strategy would be to try all possible g'.
% Therefore the best we can do is move away from the current g via a function that is trained to be insensitive to the y.

The above development leads to the network sketched in Figure~\ref{fig:network}, and
an accompanying training algorithm, \crossgrad, shown in Algorithm~\ref{alg}. Here $X,Y,D$ correspond to a minibatch of instances. Our proposed method integrates data augmentation and batch training as an alternating sequence of steps. The domain classifier is simultaneously trained with the perturbations from the label classifier network so as to be robust to label changes.  Thus, we construct cross-objectives $J_{l}$ and $J_{d}$, and update their respective parameter spaces. We found this scheme of simultaneously training both networks to be empirically superior to independent training even though the two classifiers do not share parameters.
% That is we compute both $\nabla_X J_{d}(X,D;\theta_d)$ and $\nabla_X J_{l}(X,Y;\theta_l)$ for label and domain classifiers respectively. }

%\todo{Mark where $\theta_l,\theta_d$ are involved in the diagram?}$\theta_l$ and~$\theta_d$.

%$J_{l}  =  \alpha_l J_l(x,y)  +  (1-\alpha_l)J_l(x + \epsilon\, \text{sign}(\nabla_x J_l(x,y))\nabla_x J_c(x,d),  y)$

%$J_{c}  =  \alpha_c J_c(x,d)  +  (1-\alpha_c)J_c(x + \epsilon \nabla_x J_l(x,y), d)$

%where $J_c,J_l$ represent the loss objectives of the label and domain classifier respectively.

% \todo by someone, or for someone to handle?
% SC did not put baselinestretch. Removed anyway.
%\soumen{Did you put this baselinestretch command?}
%\renewcommand{\baselinestretch}{1.2}

\begin{algorithm}[H]
  \caption{\crossgrad\ training pseudocode.}\label{alg}
  \begin{algorithmic}[1]
    \STATE {\bf Input:} Labeled data $\{(\vx_i, y_i, d_i)\}_{i=1}^M$, step sizes $\epsilon_l, \epsilon_d$, learning rate $\eta$, data augmentation weights $\alpha_l, \alpha_d$, number of training steps $n$.
    \STATE {\bf Output:} Label and domain classifier parameters $\theta_l,\theta_d$
    \STATE Initialize $\theta_l,\theta_d$
    \COMMENT {$J_l, J_d$ are loss functions for the label and domain classifiers, respectively}
    \FOR{$n$ training steps}
      \STATE Sample a labeled batch $(X,Y,D)$
%      \STATE $\zeta_l(X,Y;\theta_l) := \nabla_{X} J_l(X,Y;\theta_l)$
%      \STATE $\zeta_d(X,D;\theta_d) := \nabla_{X} J_d(X,D;\theta_d)$
      \STATE $X_d := X + \epsilon_l \cdot \nabla_{X} J_d(X, D; \theta_d)$
      \STATE $ X_l  := X + \epsilon_d \cdot \nabla_{X} J_l(X,Y;\theta_l)$
      \STATE $\theta_l \leftarrow \theta_l - \eta\nabla_{\theta_l} ((1 - \alpha_l) J_l(X, Y;\theta_l) + \alpha_l J_l(X_d, Y;\theta_l))$
      \STATE $\theta_d \leftarrow \theta_d - \eta\nabla_{\theta_d} ((1 - \alpha_d) J_d(X, D; \theta_d) + \alpha_d J_d(X_l, D; \theta_d))$
    \ENDFOR
    \renewcommand{\baselinestretch}{1}
  \end{algorithmic}
\end{algorithm}

If $y$ and $d$ are completely correlated, \crossgrad\ reduces to traditional adversarial training.  If, on the other extreme, they are perfectly uncorrelated, removing domain signal should work well.  The interesting and realistic situation is where they are only partially correlated.  \crossgrad\ is designed to handle the whole spectrum of correlations.

% \sunita{Moving the Bayesian stuff to model.tex for now.  For a possible AAAI submission, it might be a distraction.}
% \input{model}
\section{Experiments}
\label{sec:Expt}
In this section, we demonstrate that \crossgrad\ provides effective domain generalization on four different classification tasks under three different model architectures. We provide evidence that our Bayesian characterization of domains as continuous features is responsible for such generalization.  We establish that \crossgrad's domain guided perturbations provide a more consistent generalization to new domains than label adversarial perturbation ~\citep{goodfellow2014} which we denote by \goodfellow. Also, we show that DANs, a popular domain adaptation method that suppresses domain signals, provides little improvement over the baseline~\citep{ganin16,tzeng17}.

% \subsection{Datasets and setup}
We describe the four different datasets and present a summary in Table~\ref{data-summary}.

\paragraph{Character recognition across fonts.}
We created this dataset from Google Fonts\footnote{\protect\url{https://fonts.google.com/?category=Handwriting\&subset=latin}}. The task is to identify the character across different fonts as the domain.  The label set consists of twenty-six letters of the alphabet and ten numbers.  The data comprises of 109 fonts which are partitioned as  65\% train, 25\% test and 10\% validation folds.  For each font and label, two to eighteen images are generated by randomly rotating the image around the center and adding pixel-level random noise. The neural network is LeNet \citep{lenet97} modified to have three pooled convolution layers instead of two and ReLU activations instead of tanh.
% \shiv{How is it different from LeNet?}
% \sunita{ lenet is 2 pooled convolutions, followed by fully-connected layer and all activation are tanh , we use a 3 pooled conv, and have all relu activations}
%  A sample of images is shown in \ref{fig:fig1}.
% and a 2-block ResNet \citep{he16deepresidual} on a dataset of Latin fonts.

\paragraph{Handwriting recognition across authors.}
We used the LipiTk dataset that comprises of handwritten characters from the Devanagari script\footnote{\protect\url{http://lipitk.sourceforge.net/hpl-datasets.htm}}.  Each writer is considered as a domain, and the task is to recognize the character.  The images are split on writers as  60\% train, 25\% test and 15\% validation.  The neural network is the same as for the Fonts dataset above.

\paragraph{MNIST across synthetic domains.}
This dataset derived from MNIST was introduced by \cite{GhifaryBZB15}. Here, labels comprise the 10 digits and domains are created by rotating the images in multiples of 15 degrees: 0, 15, 30, 45, 60 and 75. The domains are labeled with the angle by which they are rotated, e.g., M15, M30, M45. We tested on domain M15 while training on the rest.  The network is the 2-layer convolutional one used by \cite{motiian2017CCSA}.

\paragraph{Spoken word recognition across users.}
We used the Google Speech Command Dataset\footnote{\protect\url{https://research.googleblog.com/2017/08/launching-speech-commands-dataset.html}} that consists of spoken word commands from several speakers in different acoustic settings.  Spoken words are labels and speakers are domains.  We used 20\% of domains each for testing and validation.  The number of training domains was 100 for the experiments in Table~\ref{tab:overall}.  We also report performance with varying numbers of domains later in Table~\ref{tab:speech}.
% \todo{Sid: Are we reporting the average over the ``varying numbers''? The stats are unclear here. UPDATE: ok just saw the source comments below... can we resolve this in the text?}
% We split the dataset in the ratio 60\%, 20\%, 20\% to create the train, validation and test folds.
We use the architecture of \cite{SainathP15}\footnote{ \protect\url{https://www.tensorflow.org/versions/master/tutorials/audio_recognition}}.
% \shiv{You claim the data has 1000 users but only 100 domains used during training.  Did the remaining go to test set or were the surplus training domains dropped?}
% \sunita{ To keep results fair across all different runs, we used the same validation and test set. Yes we dropped the surplus domains. I had results for full dataset, but as per Preethis suggestion removed it from table}

For all experiments, the set of domains in the training, test, and validation sets were disjoint.  We selected  hyper-parameters based on accuracy on the validation set as follows.  For  \goodfellow\ the parameter $\alpha$ was chosen from $\{0.1, 0.25, 0.75, 0.5, 0.9\}$ and for \crossgrad\ we chose $\alpha_l=\alpha_d$ from the same set of values.  We chose $\epsilon$ ranges so that $L_\infty$ norm of the perturbations are of similar sizes in \goodfellow\ and \crossgrad.  The multiples in the $\epsilon$ range came from $\{0.5, 1, 2, 2.5\}$.  The optimizer for the first three datasets is RMS prop with a learning rate ($\eta$) of 0.02 whereas for the last Speech dataset it is SGD with $\eta=0.001$ initially and 0.0001 after 15 iterations.  In \crossgrad\ networks,  $\vg$ is incorporated in the label classifier network by concatenating with the output from the last but two hidden layer. 
% domain embedding $\vg$ is concatenated with the representation layer, which is two layers before the softmax layer, of label classifier.}

% Since \goodfellow\ perturbs via $\epsilon {\text(sign)}(\nabla J)$ whereas \crossgrad\ by $\epsilon_l (\nabla J_d)$,
% So \goodfellow\ that perturbs via sign(grad) gets $\epsilon$ as the norm, while \crossgrad\ gets $\epsilon\min(norm(grad), cutoff=0.1)$ as the perturbation, and $\epsilon$ range was chosen to equate this.

\subsection{Overall comparison}
In Table~\ref{tab:overall} we compare \crossgrad\ with domain adversarial networks (\dan), label adversarial perturbation (\goodfellow), and a baseline that performs no special training.  For the MNIST dataset the baseline is CCSA~\citep{motiian2017CCSA} and D-MTAE~\citep{GhifaryBZB15}.
We observe that, for all four datasets, \crossgrad\ provides an accuracy improvement. \dan, which is designed specifically for domain adaptation, is worse than \goodfellow, which does not exploit domain signal in any way.  While the gap between \goodfellow\ and \crossgrad\ is not dramatic, it is consistent as supported by this table and other experiments that we later describe.
%\begin{minipage}{0.5\hsize}
\begin{table}
\begin{center}
\begin{tabular}{|l|l|l|} \hline
Dataset & Label  & Domain  \\ \hline
Font & 36 characters & 109 fonts \\
Handwriting & 111 characters & 74 authors \\
MNIST & 10 digits & 6 rotations \\
Speech & 12 commands & 1888 speakers  \\ \hline
\end{tabular}
\caption{\label{data-summary}Summary of datasets.}
\end{center}
\end{table}
%\end{minipage}
%\begin{minipage}{0.5\hsize}
\begin{table}[ht]
\begin{center}
\begin{tabular}{|l|r|r|r|r|} \hline
Method Name & Fonts & Handwriting  &  MNIST & Speech \\ \hline
Baseline  &  68.5 & 82.5& 95.6 & 72.6\\
\dan &  68.9 & 83.8 & 98.0& 70.4\\
\goodfellow & 71.4 & 86.3 & 97.8 &  72.7 \\
%\crossgrad without g & 87.9 &  91.8\\
\crossgrad  & \textbf{72.6} & \textbf{88.6}  & \textbf{98.6} & \textbf{73.5}\\ \hline
\end{tabular}
\end{center}
\caption{\label{tab:overall} Accuracy on four different datasets. The baseline for MNIST is CCSA.}
\end{table}
%\end{minipage}

\paragraph{Changing model architecture.}
In order to make sure that these observed trends hold across model architectures, we compare different methods with the model changed to a  2-block ResNet \citep{he16deepresidual} instead of LeNet \citep{lenet97} for the Fonts and Handwriting dataset in  Table~\ref{tab:arch}. For both datasets the ResNet model is significantly better than the LeNet model.  But even for the higher capacity ResNet model, \crossgrad\ surpasses the baseline accuracy as well as other methods like \goodfellow\ .

\begin{table}[ht]
\begin{center}
\begin{tabular}{|l|r|r|r|r|} \hline
 & \multicolumn{2}{|c|}{Fonts} & \multicolumn{2}{|c|}{Handwriting}  \\ \hline
  Method Name          & LeNet & ResNet & LeNet & ResNet \\ \hline
Baseline  &  68.5 &    80.2       & 82.5& 91.5 \\
\dan &  68.9 &      81.1         & 83.8 & 88.5 \\
\goodfellow & 71.4 &   80.5      & 86.3 & 91.8 \\
%\crossgrad without g & 87.9 &  91.8\\
\crossgrad  & \textbf{72.6} &\textbf{82.4} & \textbf{88.6}  &   \textbf{92.1}\\ \hline
\end{tabular}
\end{center}
\caption{\label{tab:arch} Accuracy with varying model architectures.}
\end{table}

% We compare \crossgrad\ against closely related training methods from the adversarial family:
% \begin{itemize}
% \item Traditional adversarial input perturbation \citep{goodfellow2014}.  Comparing against \crossgrad\ will show the extent to which \crossgrad\ can exploit relatively uncorrelated label and domain.
% \item Domain discriminative adversarial network [DAN] \citep{ganin16,tzeng17}.  Many domain adaptation papers essentially reduce to this idea.
% \end{itemize}
% We present comparisons in the context of the following applications:
% \begin{itemize}
% \item MNIST with synthetic domains.
% \item Character recognition across fonts.
% \item Handwriting recognition across authors.
% \item Isolated word recognition across speakers.
% %Face recognition across emotions.
% \end{itemize}

% \subsection{Datasets}
% - Sleep data
% - Text data
% -

\begin{figure}[th]
\begin{subfigure}[b]{0.5\textwidth}
\begin{center}
  \includegraphics[width=200pt]{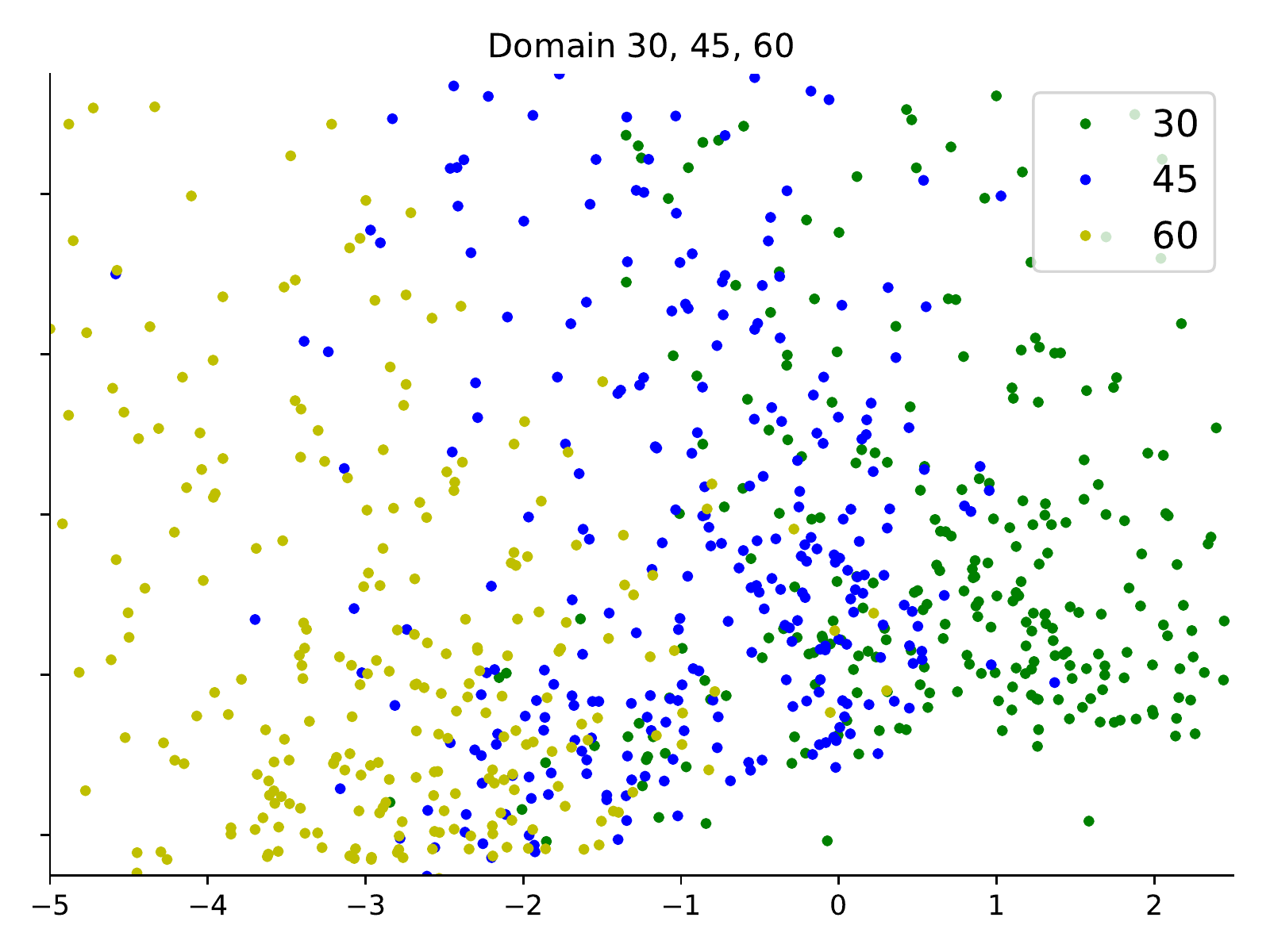}
  \caption{$\vg$ of domain \textit{45} lies between $\vg$ of \textit{60} and \textit{30}.}
  \label{fig:304560}
  \end{center}
 \end{subfigure}
 \begin{subfigure}[b]{0.5\textwidth}
\begin{center}
  \includegraphics[width=200pt]{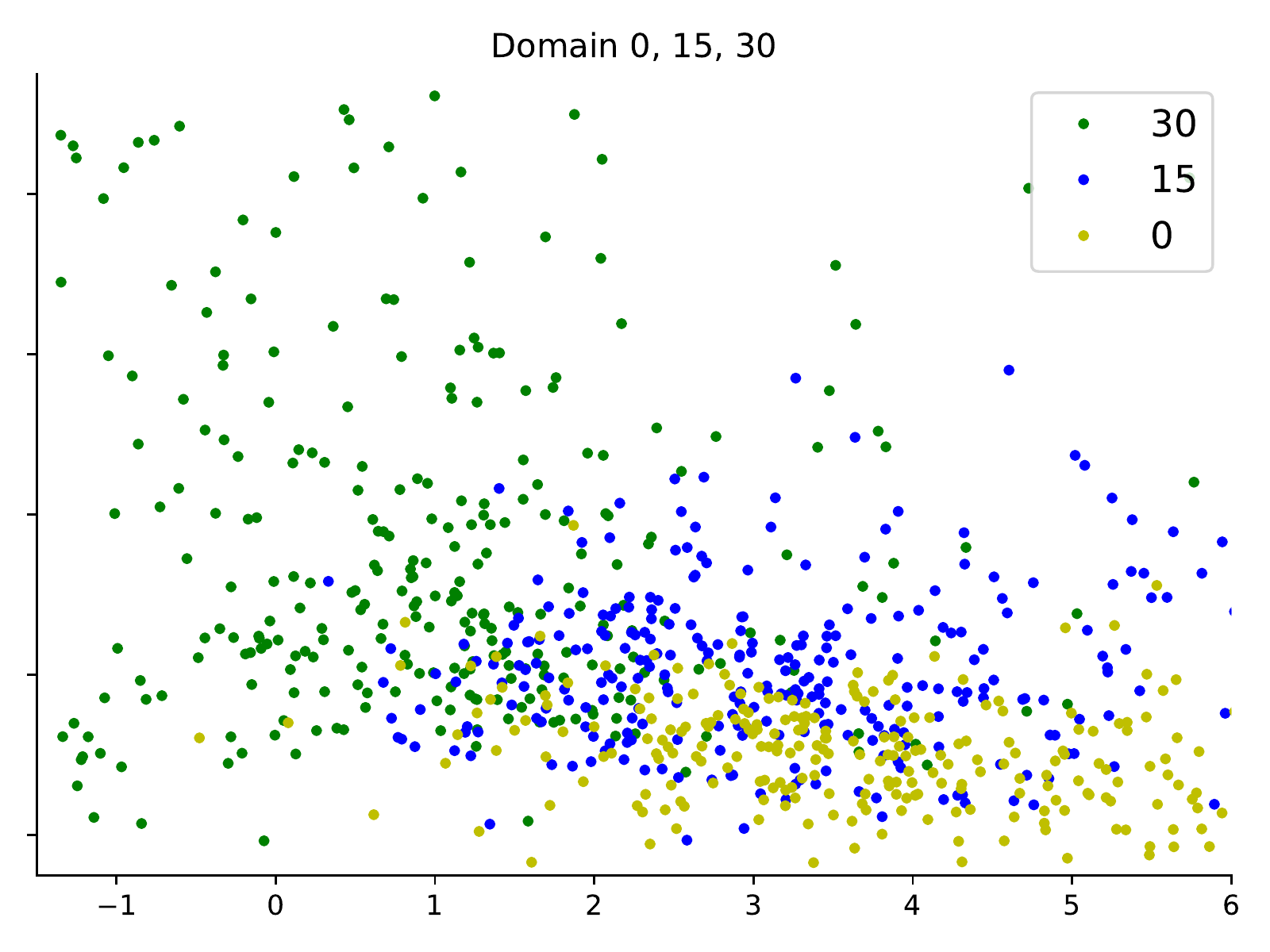}
  \caption{$\vg$ of domain \textit{15} lies between $\vg$ of \textit{0} and \textit{30}.}
  \label{fig:01530}
  \end{center}
 \end{subfigure}
 \begin{subfigure}[b]{0.5\textwidth}
 \begin{center}
  \includegraphics[width=200pt]{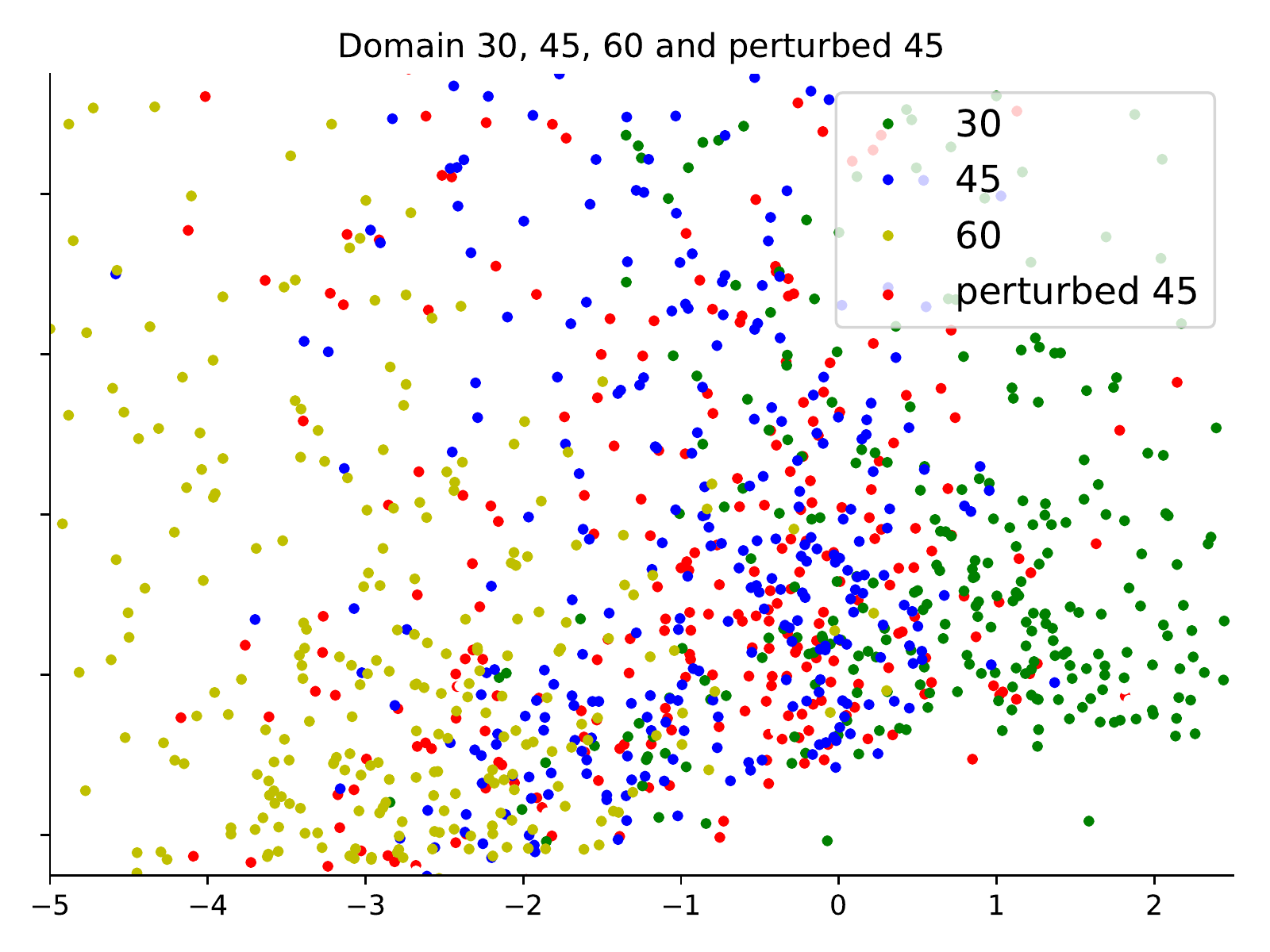}
  \caption{Adding $\vg$ of perturbed domain \textit{45} to above.}
  \label{fig:304560p}
  \end{center}
 \end{subfigure}
 \begin{subfigure}[b]{0.5\textwidth}
 \begin{center}
  \includegraphics[width=200pt]{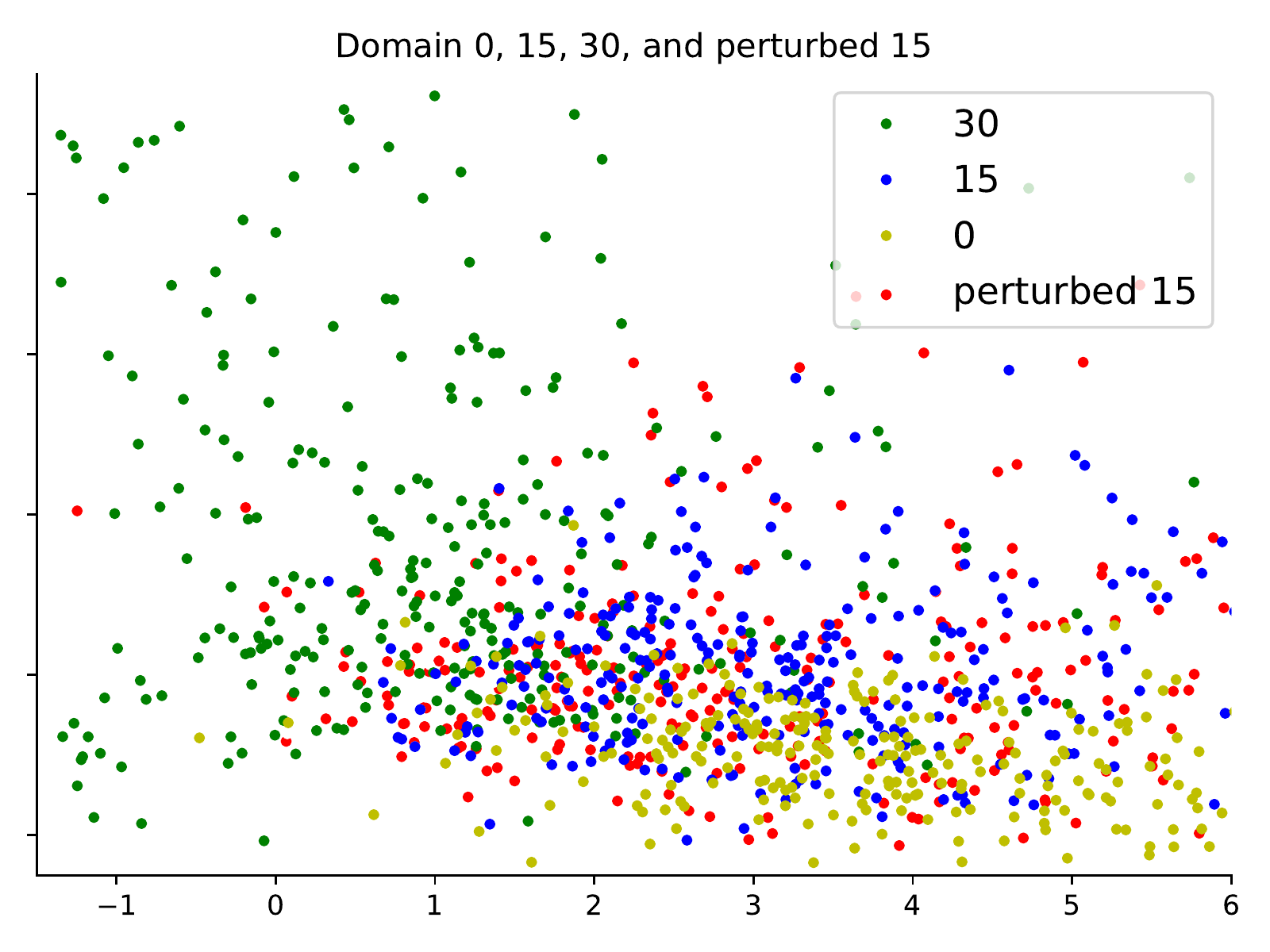}
  \caption{Adding $\vg$ of perturbed domain \textit{15} to above.}
  \label{fig:01530p}
  \end{center}
 \end{subfigure}
\caption{Comparing domain embeddings ($\vg$) across domains. Each color denotes a domain.}
\label{fig:embed_comparison}
\end{figure}

\subsection{Why does \crossgrad\ work?}

We present insights on the working of \crossgrad\ via experiments on the MNIST dataset where the domains corresponding to image rotations are easy to interpret.

In Figure~\ref{fig:304560} we show PCA projections
% \todo{Sid: I know it's pretty late for this, but how do the projections look with t-SNE or MDS?}
% \todo{Shiv: I had shared the tensorboard embeddings, if you were able to look at it. t-SNE has similar properties, but PCA looks better, and is invariant to runs }
of the $\vg$ embeddings for images from three different domains, corresponding to rotations by \textit{30, 45, 60} degrees in green, blue, and yellow respectively. The $\vg$ embeddings of domain \textit{45} (blue) lies in between the $\vg$ of domains \textit{30} (green) and \textit{60} (yellow) showing that the domain classifier has successfully extracted continuous representation of the domain even when the input domain labels are categorical. Figure~\ref{fig:01530} shows the same pattern for domains \textit{0, 15, 30}.  Here again we see that the embedding of domain \textit{15} (blue) lies in-between that of domain \textit{0}~(yellow) and \textit{30}~(green).

Next, we show that the $\vg$ perturbed along gradients of domain loss, does manage to generate images that substitute for the missing domains during training.
% The \crossgrad\ perturbations on one domain generates embeddings that are closer to the embeddings from other domains.
For example, the embeddings of the domain \textit{45}, when perturbed, scatters towards the domain \textit{30} and \textit{60} as can be seen in Figure~\ref{fig:304560p}: note the scatter of \textit{ perturbed  45} (red) points inside the \textit{30} (green) zone, without any \textit{45} (blue) points. Figure~\ref{fig:01530p} depicts a similar pattern with perturbed domain embedding points (red) scattering towards domains \textit{30} and \textit{0} more than unperturbed domain \textit{15} (blue).  For example, between x-axis -1 and 1 dominated by the green domain (domain \textit{30}) we see many more red points (perturbed domain \textit{15}) than blue points  (domain \textit{15}).  Similarly in the lower right corner of domain \textit{0} shown in yellow.
%
% \todo{Sid: This is not clear to me from the plots. Could we hand-hold the viewer through this a bit more, i.e. be very specific about what they should be focusing on when looking at the figures?}
This highlights the mechanism of \crossgrad\ working; that it is able to augment training with samples closer to unobserved domains.

\begin{figure}[tb]
\begin{subfigure}[b]{0.48\textwidth}
\begin{center}
  \includegraphics[width=210pt]{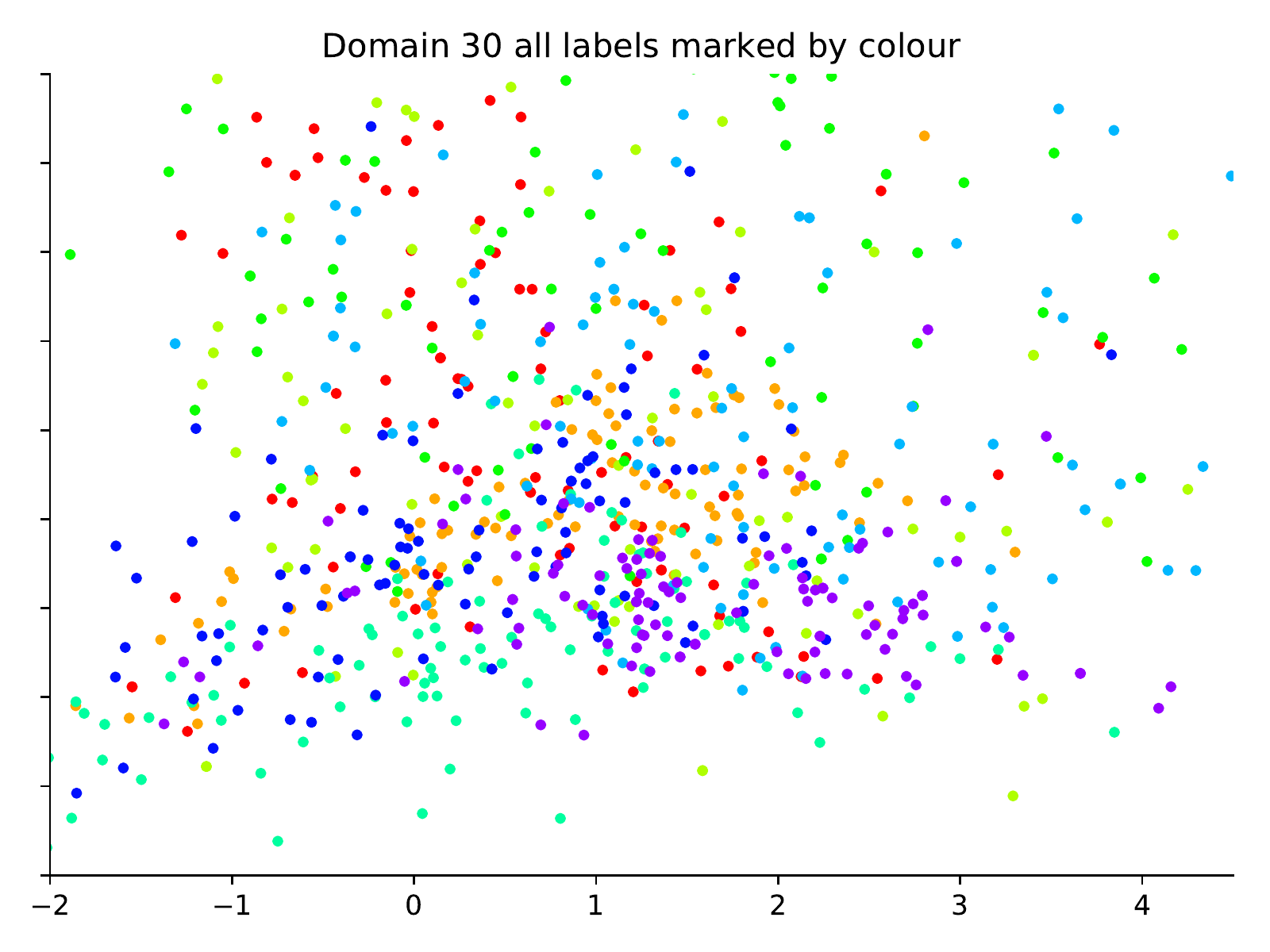}
  \caption{Domain M30}
  \label{fig:30r}
  \end{center}
 \end{subfigure}
 \begin{subfigure}[b]{0.48\textwidth}
 \begin{center}
  \includegraphics[width=210pt]{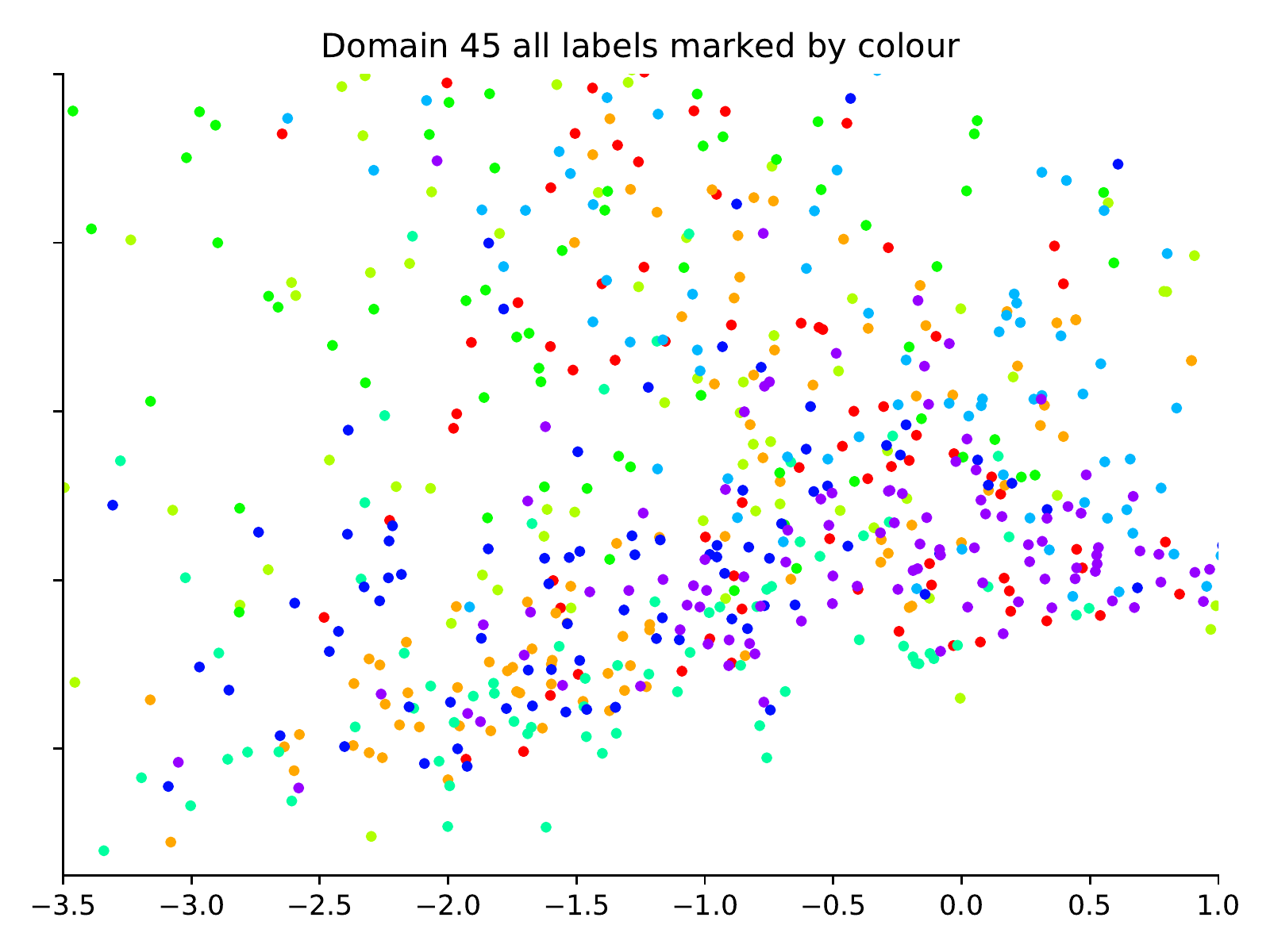}
  \caption{Domain M45}
  \label{fig:45r}
  \end{center}
 \end{subfigure}
\caption{Comparing domain embeddings ($\vg$) across labels. Each color denotes a label. Unclustered colors indicates that label information is mostly absent from $\vg$.}
\label{fig:embed_comparison_label}
\end{figure}

Finally, we observe in Figure~\ref{fig:embed_comparison_label} that the embeddings are not correlated with labels. For both domains \textit{30} and \textit{45} the colors corresponding to different labels are not clustered. This is a consequence of \crossgrad's symmetric training of the domain classifier via label-loss perturbed images.

\subsection{When is domain generalization effective?}
We next present a couple of experiments that provide insight into the settings in which \crossgrad\ is most effective.

First, we show the effect of increasing the number of training domains.  Intuitively, we expect \crossgrad\ to be most useful when training domains are scarce and do not directly cover the test domains. We verified this on the speech dataset where the number of available domains is large.  We varied the number of training domains while keeping the test and validation data fixed. Table~\ref{tab:speech} summarizes our results.  Note that \crossgrad\ outperforms the baseline and \goodfellow\ most significantly when the number of training domains is small (40). As the training data starts to cover more and more of the possible domain variations, the marginal improvement provided by \crossgrad\ decreases. In fact, when the models are trained on the full training data (consisting of more than 1000 domains), the baseline achieves an accuracy of 88.3\%, and both \crossgrad\ and \goodfellow\ provide no gains\footnote{The gap in accuracy between the baseline and \crossgrad for the case of 1000 domains is not statistically significant according to the MAPSSWE test~\citep{MAPSSWE}.} beyond that.
% at xx\% \todo{Sid: missing number} significance level as per MAPSSWE test.
DAN, like in other datasets, provides unstable gains and is difficult to tune. \goodfellow\ shows smaller relative gains than \crossgrad\ but follows the same trend of reducing gains with increasing number of domains.
%Our model generalizes better when the number of training domains is small.

\begin{table}[!t]
\begin{center}
\begin{tabular}{|l|r|r|r|r|} \hline
Method Name & 40 domains & 100 domains & 200 domains & 1000 domains \\ \hline
Baseline & 62.2 & 72.6 & 79.1 & 88.3\\ \hline
\goodfellow & +1.4 & +0.1 & -0.1 &  -0.7\\
\dan & +0.4 & -2.2 & +0.7 & -3.3\\
\crossgrad & +2.3 &  +0.9 & +0.7 & -0.4 \\ \hline
\end{tabular}
\end{center}
\caption{\label{tab:speech} Accuracy on Google Speech Command Task. For each method we show the relative improvement (positive or negative) over the baseline.}
%\todo{The numbers within brackets denote the statistical significance level as per the MAPSSWE test.}
\end{table}
%\shiv{Can you add the significance levels and the missing numbers if available?}

\iffalse
\begin{table}[!t]
\begin{center}
\begin{tabular}{|l|r|r|r|} \hline
Method Name & 20 domains & 50 domains & 100 domains  \\ \hline
Baseline & 62.2 & 72.6 & 79.1 \\
\goodfellow & 63.6 & 72.7 & 79.0   \\
\dan & 62.6 & 70.4& 79.8\\
\crossgrad & \textbf{64.5} &  \textbf{73.5} & \textbf{79.8} \\ \hline
\end{tabular}
\end{center}
\caption{\label{tab:speech} Accuracy on Google Speech Command Task.}
\end{table}
\fi

In general, how \crossgrad\ handles multidimensional, non-linear involvement of $\vg$ in determining $\vx$ is difficult to diagnose.  To initiate a basic understanding of how data augmentation supplies \crossgrad\ with hallucinated domains, we considered a restricted situation where the discrete domain is secretly a continuous 1-d space, namely, the angle of rotation in MNIST.  In this setting, a natural question is, given a set of training domains (angles), which test domains (angles) perform well under \crossgrad?

\begin{table}[ht]
\begin{center}
\begin{tabular}{|l|r|r|r|r|r|r|} \hline
Test domains $\to$ &   M0& M15 & M30 & M45  & M60 & M75 \\ \hline
\textsc{CCSA} & 84.6 & 95.6  	& 94.6 & 82.9 & 94.8 & 82.1\\
\textsc{D-MTAE} & 82.5 & 96.3  	& 93.4 & 78.6 & 94.2 & 80.5\\

\goodfellow & \textbf{89.7}  &   97.8	& 98.0 & 97.1 & 96.6& \textbf{92.1} \\
\dan & 86.7 &  98.0	& 97.8& 97.4 & 96.9 & 89.1\\
% \crossgrad without g & 86.8 &   98.1	& \textbf{98.3}& \textbf{98.0} & \textbf{97.7} \\
\crossgrad & 88.3 &   \textbf{98.6}	& \textbf{98.0} & \textbf{97.7} & \textbf{97.7} & 91.4 \\ \hline
\end{tabular}
\end{center}
\caption{\label{tab:rmnist} Accuracy on rotated MNIST.}
\end{table}

We conducted leave-one-domain-out experiments by picking one domain as the test domain, and providing the others as training domains.
In Table~\ref{tab:rmnist} we compare the accuracy of different methods. We also compare against the numbers reported by the CCSA method of domain generalization \citep{motiian2017CCSA} as reported by the authors.

It becomes immediately obvious from Table~\ref{tab:rmnist} that \crossgrad\ is beaten in only two cases: M0 and M75, which are the two extreme rotation angles.  For angles in the middle, \crossgrad\ is able to interpolate the necessary domain representation $\vg$ via `hallucination' from other training domains.  Recall from Figures~\ref{fig:304560p} and \ref{fig:01530p} that the perturbed $\vg$ during training covers for the missing test domains.  In contrast, when M0 or M75 are in the test set, \crossgrad's domain loss gradient does not point in the direction of domains outside the training domains.  If and how this insight might generalize to more dimensions or truly categorical domains is left as an open question.

% Note that while our method is better than \textsc{CCSA} and \dan\ on all experiments, adversarial perturbations do better than \crossgrad\ on M0 and M75. This is not surprising, as our method is designed to do better when the unseen domain can be interpolated on training domains.

% \subsection{Explore time-series activity recognition data}

% \subsection{Movie reviews}
% We test our method on \todo{name} different textual data. \textbf{aclIMDB} is a sentiment classification task on movie reviews collected from IMDB by \cite{Maas11} .

% \begin{table}[!htb]
% \begin{center}
% \begin{tabular}{|l|r|r|r|} \hline
% Name &  Classes & Train & Test \\ \hline
% aclIMDB & 2 & 25000 &  25000 \\
% %%
% \end{tabular}
% \end{center}
% \caption{\label{textdata-summary} Summary statistics of the text datasets}

% \end{table}

% \subsection{Ablation experiments}
% Can we show that even if we could separate domain information, it is better to retain it for our goal of multi-domain adaptation?

\section{Conclusion}
\label{sec:End}

Domain $d$ and label $y$ interact in complicated ways to influence the observable input $\vx$.  Most domain adaption strategies implicitly consider the domain signal to be extraneous and seek to remove its effect to train more robust label predictors.  We presented \crossgrad, which considers them in a more symmetric manner.  \crossgrad\ provides a new data augmentation scheme based on the $y$ (respectively, $d$) predictor using the gradient of the $d$ (respectively, $y$) predictor over the input space, to generate perturbations.  Experiments comparing \crossgrad\ with various recent adversarial paradigms show that \crossgrad\ can make better use of partially correlated $y$ and $d$, without requiring explicit distributional assumptions about how they affect~$\vx$.  \crossgrad\ is at its best when training domains are scarce and do not directly cover test domains well.  Future work includes extending  \crossgrad\ to exploit labeled or unlabeled data in the test domain, and integrating the best of \goodfellow\ and \crossgrad\ into a single algorithm.

\subsubsection*{Acknowledgements}
We gratefully acknowledge the support of NVIDIA Corporation with the donation of Titan X GPUs used for this research.
We thank Google for supporting travel to the conference venue.
\bibliographystyle{iclr2018_conference}
\bibliography{sample}

\begin{thebibliography}{34}
\providecommand{\natexlab}[1]{#1}
\providecommand{\url}[1]{\texttt{#1}}
\expandafter\ifx\csname urlstyle\endcsname\relax
  \providecommand{\doi}[1]{doi: #1}\else
  \providecommand{\doi}{doi: \begingroup \urlstyle{rm}\Url}\fi

\bibitem[Balestrino et~al.(1984)Balestrino, Maria, and
  Sciavicco]{BalestrinoMS84}
A.~Balestrino, G.~De Maria, and L.~Sciavicco.
\newblock Robust control of robotic manipulators.
\newblock \emph{IFAC Proceedings Volumes}, 17\penalty0 (2):\penalty0
  2435--2440, 1984.

\bibitem[Gan et~al.(2016)Gan, Yang, and Gong]{GanYG16}
C~Gan, T.~Yang, and B.~Gong.
\newblock Learning attributes equals multi-source domain generalization.
\newblock In \emph{CVPR}, pp.\  87--97, 2016.

\bibitem[Ganin et~al.(2016)Ganin, Ustinova, Ajakan, Germain, Larochelle,
  Laviolette, Marchand, and Lempitsky]{ganin16}
Y.~Ganin, E.~Ustinova, H.~Ajakan, P.~Germain, H.~Larochelle, F.~Laviolette,
  M.~Marchand, and V.~S. Lempitsky.
\newblock Domain-adversarial training of neural networks.
\newblock \emph{JMLR}, 17:\penalty0 59:1--59:35, 2016.

\bibitem[Ghifary et~al.(2015)Ghifary, Bastiaan~Kleijn, Zhang, and
  Balduzzi]{GhifaryBZB15}
M.~Ghifary, W.~Bastiaan~Kleijn, M.~Zhang, and D.~Balduzzi.
\newblock Domain generalization for object recognition with multi-task
  autoencoders.
\newblock In \emph{ICCV}, pp.\  2551--2559, 2015.

\bibitem[Gillick \& Cox(1989)Gillick and Cox]{MAPSSWE}
L.~Gillick and S.~J. Cox.
\newblock Some statistical issues in the comparison of speech recognition
  algorithms.
\newblock In \emph{ICASSP}, pp.\  532--535, 1989.

\bibitem[Gong et~al.(2012)Gong, Shi, Sha, and Grauman]{Gong2012}
B.~Gong, Y.~Shi, F.~Sha, and K.~Grauman.
\newblock Geodesic flow kernel for unsupervised domain adaptation.
\newblock In \emph{CVPR}, pp.\  2066--2073, 2012.

\bibitem[Goodfellow et~al.(2014)Goodfellow, Shlens, and
  Szegedy]{goodfellow2014}
I.~Goodfellow, J.~Shlens, and C.~Szegedy.
\newblock Explaining and harnessing adversarial examples.
\newblock \emph{arXiv preprint arXiv:1412.6572}, 2014.

\bibitem[Gopalan et~al.(2011)Gopalan, Li, Ruonan, and Chellappa]{Gopalan2011}
R.~Gopalan, R.~Li, Ruonan, and R.~Chellappa.
\newblock Domain adaptation for object recognition: An unsupervised approach.
\newblock In \emph{ICCV}, pp.\  999--1006, 2011.

\bibitem[He et~al.(2016)He, Zhang, Ren, and Sun]{he16deepresidual}
K.~He, X.~Zhang, S.~Ren, and J.~Sun.
\newblock Deep residual learning for image recognition.
\newblock In \emph{CVPR}, pp.\  770--778, 2016.

\bibitem[Huang \& Belongie(2017)Huang and Belongie]{HuangB17}
X.~Huang and S.~J. Belongie.
\newblock Arbitrary style transfer in real-time with adaptive instance
  normalization.
\newblock \emph{CoRR}, abs/1703.06868, 2017.

\bibitem[III(2007)]{Daume2007}
H.~Daum{\'e} III.
\newblock Frustratingly easy domain adaptation.
\newblock pp.\  256--263, 2007.

\bibitem[III et~al.(2010)III, Kumar, and Saha]{Kumar2010}
H.~Daum{\'e} III, A.~Kumar, and A.~Saha.
\newblock Co-regularization based semi-supervised domain adaptation.
\newblock In \emph{NIPS}, pp.\  478--486, 2010.

\bibitem[Jiang \& Zhai(2007)Jiang and Zhai]{Jiang2007}
J.~Jiang and C.~Zhai.
\newblock Instance weighting for domain adaptation in {NLP}.
\newblock In \emph{ACL}, pp.\  264--271, 2007.

\bibitem[Khosla et~al.(2012)Khosla, Zhou, Malisiewicz, Efros, and
  Torralba]{ECCV12_Khosla}
A.~Khosla, T.~Zhou, T.~Malisiewicz, A.~Efros, and A.~Torralba.
\newblock Undoing the damage of dataset bias.
\newblock In \emph{ECCV}, pp.\  158--171, 2012.

\bibitem[Kuan{-}Chuan et~al.(2017)Kuan{-}Chuan, Ziyan, and Ernst]{PengWE17}
P.~Kuan{-}Chuan, W.~Ziyan, and J.~Ernst.
\newblock Zero-shot deep domain adaptation.
\newblock \emph{CoRR}, abs/1707.01922, 2017.

\bibitem[LeCun et~al.(1998)LeCun, Bottou, Bengio, and Haffner]{lenet97}
Y.~LeCun, L.~Bottou, Y.~Bengio, and P.~Haffner.
\newblock Gradient-based learning applied to document recognition.
\newblock \emph{Proceedings of the IEEE}, 86\penalty0 (11):\penalty0
  2278--2324, 1998.

\bibitem[Li et~al.(2013)Li, Zhang, Wang, Cao, Shamir, and Cohen-Or]{Li13}
H.~Li, H.~Zhang, Y.~Wang, J.~Cao, A.~Shamir, and D.~Cohen-Or.
\newblock Curve style analysis in a set of shapes.
\newblock \emph{Computer Graphics Forum}, 32\penalty0 (6), 2013.

\bibitem[Mansour et~al.(2009)Mansour, Mohri, and Rostamizadeh]{Mansour2009}
Y.~Mansour, M.~Mohri, and A.~Rostamizadeh.
\newblock Domain adaptation with multiple sources.
\newblock In \emph{NIPS}, pp.\  1041--1048, 2009.

\bibitem[Miyato et~al.(2016)Miyato, Dai, and Goodfellow]{MiyatoDG16}
T.~Miyato, A.~Dai, and I.~Goodfellow.
\newblock Virtual adversarial training for semi-supervised text classification.
\newblock \emph{CoRR}, abs/1605.07725, 2016.

\bibitem[Motiian et~al.(2017)Motiian, Piccirilli, Adjeroh, and
  Doretto]{motiian2017CCSA}
S.~Motiian, M.~Piccirilli, D.~A. Adjeroh, and G.~Doretto.
\newblock Unified deep supervised domain adaptation and generalization.
\newblock In \emph{ICCV}, pp.\  5715--5725, 2017.

\bibitem[Muandet et~al.(2013)Muandet, Balduzzi, and Schölkopf]{MuandetBS13}
K.~Muandet, D.~Balduzzi, and B.~Schölkopf.
\newblock Domain generalization via invariant feature representation.
\newblock In \emph{ICML}, pp.\  10--18, 2013.

\bibitem[Saenko et~al.(2010)Saenko, Kulis, Fritz, and Darrell]{Saenko2010}
K.~Saenko, B.~Kulis, M.~Fritz, and T.~Darrell.
\newblock Adapting visual category models to new domains.
\newblock \emph{ECCV}, pp.\  213--226, 2010.

\bibitem[Sainath \& Parada(2015)Sainath and Parada]{SainathP15}
T.~Sainath and C.~Parada.
\newblock Convolutional neural networks for small-footprint keyword spotting.
\newblock In \emph{INTERSPEECH}, pp.\  1478--1482, 2015.

\bibitem[Saon et~al.(2013)Saon, Soltau, Nahamoo, and Picheny]{Saon2013}
G.~Saon, H.~Soltau, D.~Nahamoo, and M.~Picheny.
\newblock Speaker adaptation of neural network acoustic models using i-vectors.
\newblock In \emph{ASRU}, pp.\  55--59, 2013.

\bibitem[Szegedy et~al.(2013)Szegedy, Zaremba, Sutskever, Bruna, Erhan,
  Goodfellow, and Fergus]{szegedy14}
C.~Szegedy, W.~Zaremba, I.~Sutskever, J.~Bruna, D.~Erhan, I.~Goodfellow, and
  R.~Fergus.
\newblock Intriguing properties of neural networks.
\newblock \emph{arXiv preprint arXiv:1312.6199}, 2013.

\bibitem[Tzeng et~al.(2015)Tzeng, Hoffman, Darrell, and Saenko]{tzeng15}
E.~Tzeng, J.~Hoffman, T.~Darrell, and K.~Saenko.
\newblock Simultaneous deep transfer across domains and tasks.
\newblock In \emph{ICCV}, pp.\  4068--4076, 2015.

\bibitem[Tzeng et~al.(2017)Tzeng, Hoffman, Darrell, and Saenko]{tzeng17}
E.~Tzeng, J.~Hoffman, T.~Darrell, and K.~Saenko.
\newblock Adversarial discriminative domain adaptation.
\newblock In \emph{CVPR}, pp.\  2962--2971, 2017.

\bibitem[Upchurch et~al.(2016)Upchurch, Snavely, and Bala]{UpchurchSB16}
P.~Upchurch, N.~Snavely, and K.~Bala.
\newblock From {A} to {Z:} supervised transfer of style and content using deep
  neural network generators.
\newblock \emph{CoRR}, abs/1603.02003, 2016.

\bibitem[Wolovich \& Elliott(1984)Wolovich and Elliott]{WolovichE84}
W.~A. Wolovich and H.~Elliott.
\newblock A computational technique for inverse kinematics.
\newblock In \emph{IEEE Conference on Decision and Control}, pp.\  1359--1363,
  1984.

\bibitem[Woodland(2001)]{Woodland2001}
P.~C. Woodland.
\newblock Speaker adaptation for continuous density {HMMs}: A review.
\newblock In \emph{ITRW}, pp.\  11--19, 2001.

\bibitem[Xu et~al.(2014)Xu, Li, Niu, and Xu]{XuLNX14}
Z.~Xu, W.~Li, L.~Niu, and D.~Xu.
\newblock Exploiting low-rank structure from latent domains for domain
  generalization.
\newblock In \emph{ECCV}, pp.\  628--643, 2014.

\bibitem[Yang \& Hospedales(2015)Yang and Hospedales]{Yang2015}
Y.~Yang and T.~M. Hospedales.
\newblock A unified perspective on multi-domain and multi-task learning.
\newblock In \emph{ICLR}, 2015.

\bibitem[Yang \& Hospedales(2016)Yang and Hospedales]{Yang2016}
Y.~Yang and T.~M. Hospedales.
\newblock Multivariate regression on the {G}rassmannian for predicting novel
  domains.
\newblock In \emph{CVPR}, pp.\  5071--5080, 2016.

\bibitem[Zhang et~al.(2015)Zhang, Gong, and Sch{\"o}lkopf]{Zhang2015}
K.~Zhang, M.~Gong, and B.~Sch{\"o}lkopf.
\newblock Multi-source domain adaptation: A causal view.
\newblock In \emph{AAAI}, pp.\  3150--3157, 2015.

\end{thebibliography}

\section*{Appendix}

\mypara{Relating the ``natural'' perturbations of $\vx$ and $\vgmap$.} In Section \ref{sec:probmodel}, we claimed that the intuitive perturbation $\epsilon \nabla_{\vx_i} J_{d}(\vx_i, d_i)$ of $\vx_i$ attempts to induce the intuitive perturbation $\epsilon' \nabla_{\vgmap_i} J_{d}(\vx_i, d_i)$ of $\vgmap_i$, even though the exact relation between a perturbation of $\vx_i$ and that of $\vgmap_i$ requires the metric tensor transpose $\jac \jac^\top$. We will now prove this assertion. Of course, an isometric map $G : \vx \mapsto \vg$, with an orthogonal Jacobian ($\jac^{-1} = \jac^\top$), trivially has this property, but we present an alternative derivation which may give further insight into the interaction between the perturbations in the general case.

Consider perturbing $\vgmap_i$ to produce $\vg'_i = \vgmap_i + \epsilon' \nabla_{\vgmap_i} J_{d}(\vx_i, d_i)$. This yields a new augmented input instance $\vx'_i$ as
\begin{align*}
G^{-1}(\vg'_i) = G^{-1}\bigl(G(\vx_i) +
\epsilon' \nabla_{\vgmap_i} J_{d}(\vx_i, d_i)\bigr).
\end{align*}
We show next that the perturbed  $\vx_i'$ can be approximated by $\vx_i + \epsilon \nabla_{\vx_i} J_d(\vx_i, d_i)$.

%We show that when  $G(\vx)=\vgmap$ and $\vgmap$ is perturbed by $\Delta \vgmap = \nabla_{\vgmap} J_d(\vx, d)$, then change in $\vx$ reduces to $\epsilon' \nabla_\vx J_d(\vx, d)$.

In this proof we drop the subscript $i$ for ease of notation.
In the forward direction, the relationship between $\Delta \vx$ and $\Delta \vgmap$ can be expressed using the Jacobian $\jac$ of $\vgmap$ w.r.t. $\vx$:
\begin{align*}
\Delta \vgmap = \jac \Delta \vx
\end{align*}
To invert the relationship for a non-square and possibly low-rank Jacobian, we use the Jacobian transpose method devised for inverse kinematics~\citep{BalestrinoMS84,WolovichE84}. Specifically, we write $\Delta \vgmap = \vgmap(\vx') - \vgmap(\vx)$, and recast the problem as trying to minimize the squared L2 error
\begin{align*}
L_{\vgmap}(\vx) = \tfrac{1}{2} \left( \vgmap(\vx') - \vgmap(\vx) \right)^\top \left( \vgmap(\vx') - \vgmap(\vx) \right)
\end{align*}
with gradient descent. The gradient of the above expression w.r.t. $\vx$ is
\begin{eqnarray*}
%\nabla_{\vx} L_{\vgmap} & = & \nabla_{\vx} \left( \frac{1}{2} \left( \vgmap(\vx') - \vgmap(\vx) \right)^T \left( \vgmap(\vx') - \vgmap(\vx) \right) \right) \\
\nabla_{\vx} L_{\vgmap} & = & -\left( \Delta \vgmap^\top \ \frac{\partial \vgmap}{\partial \vx} \right)^\top \ = \ -\jac^\top \Delta \vgmap
\end{eqnarray*}
Hence, the initial gradient descent step to affect a change of $\Delta \vgmap$ in the domain features would increment $\vx$ by $\epsilon \jac^\top \Delta \vgmap$. The Jacobian, which is a matrix of first partial derivatives, can be computed by back-propagation.  Thus we get
\iffalse
If \shiv{on the next line where the grad is explicitly written in component wise derivative}
$\Delta g = \nabla_{\vgmap} J_d(\vgmap, d)$, then
\fi
%
\begin{align*}
\Delta \vx = \epsilon \jac^\top \nabla_{\vgmap} J_d(\vgmap, d)
= \epsilon \sum_{i} \frac{\partial \vgmapcoord^i}{\partial \vx}^\top \frac{\partial J_d(\vgmap, d)}{\partial \vgmapcoord^i },
\end{align*}
which, by the chain rule, gives
\begin{align*}
\Delta \vx = \epsilon \nabla_{\vx} J_d(\vx, d).
\end{align*}
% Assuming local linearity and for a suitable $\epsilon$, we can approximate the complete gradient descent process with this single step.

\end{document}